\documentclass[11pt,a4paper]{article}
\usepackage[hyperref]{mobivsr}
\usepackage{times}
\usepackage{latexsym}
\usepackage{multicol}
\usepackage{amsmath}
\usepackage{url}
\usepackage{graphicx}
\usepackage{booktabs}
\usepackage{comment}

\aclfinalcopy % Uncomment this line for the final submission
\title{MobiVSR: A Visual Speech Recognition Solution for Mobile Devices}

\author{Nilay Shrivastava\thanks{\hspace{0.2cm}Equal contribution} \\
NSIT-Delhi \\nilays.co@nsit.net.in \\\And Astitwa Saxena\footnotemark[1] \\  NSIT-Delhi \\  astitwas.co@nsit.net.in \\\And   Yaman Kumar\footnotemark[1] \\ Adobe Systems, Noida \\  ykumar@adobe.com \\\AND Rajiv Ratn Shah \\ IIIT-Delhi \\ rajivratn@iiitd.ac.in \\\And   Debanjan Mahata \\ Bloomberg LP \\ dmahata@bloomberg.net \\\And  Amanda Stent\\  Bloomberg LP \\ astent@bloomberg.net \\}

\begin{document}
\maketitle

\begin{abstract}

Visual speech recognition (VSR) is the task of recognizing spoken language from video input only, without any audio. VSR has many applications as an assistive technology, especially if it could be deployed in mobile devices and embedded systems. The need for intensive computational resources and large memory footprint are two of the major obstacles in developing neural network models for VSR in a resource constrained environment. We propose a novel end-to-end deep neural network architecture for word level VSR called MobiVSR with a design parameter that aids in balancing the model's accuracy and parameter count. We use depthwise-separable 3D convolution for the first time in the domain of VSR and show how it makes our model efficient. MobiVSR achieves an accuracy of 73\% on a challenging Lip Reading in the Wild dataset with 6 times fewer parameters and 20 times lesser memory footprint than the current state of the art. MobiVSR can also be compressed to 6 MB by applying post training quantization.
\end{abstract}
% AS 3/4 - Fix the Yx in the abstract

% AS - if you want to you might consider INTERSPEECH as an alternative publication venue for this work. INTERSPEECH is more speechy/signal processy than ACL and also more aware of these hardware and compute limitations.

\section{Introduction}
\label{section:intro}
Visual speech recognition (VSR) is the task of recognizing spoken language from video input only, without any audio. Similar to ASR (audio speech recognition), VSR has a multitude of applications. In general, VSR can be used to augment/replace audio speech recognition for situations where speech cannot be heard or produced. For example, if a person has a laryngectomy or voice-box cancer, dysarthria or in a situation where one needs to understand a speaker in a noisy environment. %<ISSUE should we add a line statubg VSR is not deployable>?
The specific aim of this work is to make VSR technology deployable, especially in mobile environments (such as cars) and on hand-held devices (as an assistive technology). 
% YK-ARA
However the broader aim of this work is to provide a new deep learning construct to optimize deep learning frameworks for video classification tasks. We show this by taking lipreading as a language recognition problem on top of a video classification task. %<ISSUE this line is slighty not understandable>
% YK-ARA 
% 1.	When Speech is missing – 
% a.	In case of voice box cancer patients (laryngectomees, laryngeal cancer, etc), it becomes highly difficult for them to speak anything. Thus, an AVSR application can be a boon for them in helping them to speak.
% b.	In cases where speech is missing either intentionally or non-intentionally. Examples include silent movies, cases when microphones are broken, CCTV footages, etc
% 2.	When speech is noisy
% a.	Examples include traffic conditions while driving. With more and more cars now carrying artificially intelligent assistants which help in various tasks ranging from navigation to calling, interacting with these assistants is a pretty hard task. This becomes frustratingly unmanageable especially since both the hands of a driver are busy in driving. Here video modality can be remarkably helpful.
% b.	Video conferencing where the speech might be noisy due to a number of reasons.

Recent research in VSR has focused primarily on either recognizing language \cite{chung2016lip, chung2017lip}, or reconstructing it \cite{kumar2018mylipper, kumar2018harnessing}. The application of deep learning techniques has produced models that perform substantially better on lip reading datasets than earlier methods \cite{petridis2017end, resnet}. However, the major challenge for all these models is to overcome the unique challenges presented at the time of actual use or deployment. 

For example, the state-of-the-art model for word level VSR \cite{lipres} uses a novel architecture that incorporates a 3D convolution kernel as a front-end to extract features from the video stream and a residual architecture \cite{resnet} on top of it for predicting the word spoken. The architecture has more than 25 million parameters, occupies 130 MB of disk space and involves 290 million FLOPs for inference. Memory and computation of this order is prohibitively expensive for mobile devices. For instance, the Apple store places a hard limit of 150 MBs on a fully functional app that should include all its components. Furthermore, according to empirical observations made on iOS \cite{iosApps}, an app taking more than 50\% of the total RAM available at runtime, often crashes. 

In addition, performing inference using such models require significant amounts of energy due to memory access and floating point arithmetic \cite{energystat}. Table \ref{table:energy} shows energy consumption per operation on an Intel 45nm based system; accessing DRAM consumes $\approx2500$ times more energy than floating point addition and therefore dominates energy expenditure. Memory access demands depend on the number of parameters and the intermediate results generated during a forward pass of the neural network, both of which are quite high in all VSR models (Table \ref{table:results}). This means performing inference on VSR model consumes a substantial amount of energy. Therefore, battery drain is a significant issue with such models. %<ISSUE intro mein a sentence about CO2>
\begin{table*}[]
\centering
\begin{tabular}{@{}lllll@{}}
\toprule
\multicolumn{1}{c}{\textbf{Operation}} & \multicolumn{1}{c}{\textbf{Bit width}} & \multicolumn{1}{c}{\textbf{Energy (pJ)}} & \multicolumn{2}{c}{\textbf{\begin{tabular}[c]{@{}c@{}}Relative Energy \\ consumed per bit\end{tabular}}} \\ \midrule
float add & 32 & 0.9 & \multicolumn{2}{l}{1} \\
float multiply & 32 & 3.7 & \multicolumn{2}{l}{4.11} \\
DRAM access & 64 & 1300-2600 & \multicolumn{2}{l}{722.22 - 1444.44} \\ \bottomrule
\end{tabular}
\caption{\label{table:energy}Energy expenditure for different operations in 45nm technology \cite{energystat}}
\end{table*}
Thus, lipreading models, in general, have prohibitively large memory and energy requirements while also making their response-times unacceptable for real-time applications. In this paper, we try to optimize on all four fronts, \textit{i.e.}, memory, time, size and energy while keeping the performance  stable. We achieve this using depthwise separable 3D convolution, which we introduce, for the first time, in video classification domain.

% Interestingly, Apple imposes a size limit of 150 MB on apps downloaded on cellular data for iOS .. can this be included in here somehow?
% YK - Added to this is the fact that runtime of an App cannot exceed approx 50% of the RAM (https://stackoverflow.com/questions/5887248/ios-app-maximum-memory-budget) on iOS. These limits are on iOS, consider more low end devices and you would be amazed.

% YK - Cite https://stackoverflow.com/questions/5887248/ios-app-maximum-memory-budget at XXX

% \begin{table}[htbp]
% \scalebox{0.80}{\begin{tabular}{@{}lllll@{}}
% \toprule
% \multicolumn{1}{c}{\textbf{Operation}} & \multicolumn{1}{c}{\textbf{Bit width}} & \multicolumn{1}{c}{\textbf{Energy (pJ)}} & \multicolumn{2}{c}{\textbf{\begin{tabular}[c]{@{}c@{}}Relative Energy \\ consumed per bit\end{tabular}}} \\ \midrule
% float add & 32 & 0.9 & \multicolumn{2}{l}{1} \\
% float multiply & 32 & 3.7 & \multicolumn{2}{l}{4.11} \\
% DRAM access & 64 & 1300-2600 & \multicolumn{2}{l}{722.22 - 1444.44} \\ \bottomrule
% \end{tabular}}
% \caption{\label{table:energy}Energy expenditure for different operations in 45nm technology.}
% \end{table}
Thus, the main contributions we make in this paper are:

\noindent \textbf{$\bullet$} We \textbf{generalize 3D convolutions} by using depthwise-separable convolutions and show the applicability of this technique for \textit{the first time} in the video classification domain. This technique helps us to reduce the parameter count and computational complexity vis-a-vis standard convolution (Section \ref{section:architecture}). This, in turn helps us to achieve better runtime and memory while keeping the accuracy competent. %design parameter aspect

\noindent \textbf{$\bullet$} We try to provide a new, atomic deep learning construct containing all the optimizations we make, to deep learning practitioners. They can also use this to optimize their models. We call this construct a \textbf{LipRes block}. This block also additionally serves as a memory-performance trade-off handle. In the paper, it is modeled as a design hyperparameter, $\alpha$ which allows practitioners to trade off accuracy and model size depending on the use case and constraints. %<ISSUE Lipres isnt modeled by alpha.. alpha the bas number control kar raha hai>
It should be noted that the techniques used in MobiVSR are independent from and complement model compression techniques \cite{quantization, xnornet, tensorfactor, modelcompression}.

\noindent \textbf{$\bullet$} We present the MobiVSR architecture, which for the first time, addresses the problem of deploying visual speech recognition models on \textbf{resource constrained devices} (Section \ref{section:architecture}).

\noindent \textbf{$\bullet$} We show that MobiVSR achieves accuracy of 73\% on lipreading in the wild task, in spite of having \textbf{$6 \times$ fewer parameters and $20 \times$ smaller model size than the state-of-the-art model}. Using additional parameter quantization techniques, MobiVSR's size can be reduced to \textbf{6MB} (Appendix Section \ref{memory_access}).

%NTC
\noindent \textbf{$\bullet$} MobiVSR has been made keeping the following perspectives in mind: first, deployment on mobile platforms and two, general optimization of video recognition architectures. Thus, we perform \textbf{energy and environmental cost modelling and comparison} for MobiVSR and all other models (Section \ref{section:quant}). Comparisons reveal that MobiVSR has \textit{2.9 times lesser} energy impact than other conventional models.

\section{Related Work}

% AS - think you need one paragraph here about the research history of lip reading
Research on lip reading spans across centuries \cite{deafanddumb}. Several psychological studies have demonstrated that lower level visual information helps in hearing \cite{demorest1991computational, dodd1987hearing}. Experiments and research studies have shown that people with \cite{bernstein1998makes, marschark1998mouth}, and without \cite{summerfield1992lipreading}, hearing impairment use visual cues for augmenting their understanding of what a speaker is trying to say. As noted by \cite{chen1998audio}, skilled lip readers look at the configuration and movement of tongue, lips and teeth.

In computational approaches, lip reading is considered as a classification task where the input is a silent video of a speaker utterance and the output is to predict that utterance. Mostly, words or phrases are identified and selected from a limited lexicon, for example just digits \cite{chen1998audio, pachoud2008macro, sui2015listening}. Early approaches use feature engineering and trains classification models using them \cite{ngiam2011multimodal, noda2015audio}. \cite{cornu2015reconstructing}, use hand-engineered features to reconstruct audio from video, using a deep-learning network. This method then was modified by \cite{ephrat2017improved} who use a CNN over the entire face of the speaker.
Several end-to-end deep learning models have also been developed which rely on a combination of CNN and RNN \cite{assael2016lipnet, wand2016lipreading}.

Until recently datasets for lipreading were limited by having training sets capturing only a single view of the speakers, and having words from a restricted vocabulary. Recently datasets have become available that contain multiple views \cite{petridis2017end, kumar2018mylipper, kumar2018harnessing}. \cite{chung2016lip} developed a large scale dataset for lip reading with hundreds of distinct words, thousands of instances for each word, and over a thousand speakers. We use this dataset for evaluating MobiVSR (Section \ref{data}).

% AS - with respect to the below, there has been work on minimal parameter models for speech recognition
% starting at least with Pocket Sphinx and including eg https://ieeexplore.ieee.org/abstract/document/8639603/, https://ieeexplore.ieee.org/abstract/document/7472823/, Mori, T., Tjandra, A., Sakti, S., & Nakamura, S. (2018). Compressing End-to-end ASR Networks by Tensor-Train Decomposition. Proc. Interspeech 2018, 806-810., https://arxiv.org/pdf/1806.01248, http://papers.nips.cc/paper/8261-fully-neural-network-based-speech-recognition-on-mobile-and-embedded-devices.pdf, etc.

Very little work on VSR has focused on developing efficient architectures; however, there is work on this task in image classification and in ASR. The problem of efficient architecture development for image classification was introduced in \cite{squeezenet}; SqueezeNet achieves accuracy on par with AlexNet \cite{alexnet} but uses far fewer parameters, by using convolutional blocks having $3 \times 3$ convolutions followed by $1 \times 1$ instead of the large $5 \times 5$ kernel used in AlexNet.
MobileNet \cite{mobilenet1} uses depthwise-separable convolution \cite{firstdepth,seconddepth}, for parameter reduction. The MobileNet-V2 architecture \cite{mobilenet2} improves MobileNet by adding residual connections within the MobileNet depthwise-separable modules. This idea was a major influence in the design of MobiVSR (Section \ref{section:architecture}).

In the domain of Automatic Speech Recognition (ASR) from audio, a major contribution was PocketSphinx \cite{huggins2006pocketsphinx}, a large vocabulary, speaker-independent continuous speech recognition engine suitable for hand-held devices. More recently, different architectures of compressed RNNs have been proposed for ASR \cite{prabhavalkar2016compression, mori2018compressing, zhang2018dynamically}. \cite{park2018fully} constructs an acoustic model by combining simple recurrent units (SRUs) and depth-wise 1-dimensional convolution layers for multi time step parallelization; this results in reductions in DRAM access and increase in processing speed, allowing real-time on-device ASR on mobile and embedded devices. 
% Efforts have been made to investigate whether on-device ASR can be used for a virtual reading companion using recordings obtained from children both in a controlled environment and in the field \cite{loukina2018evaluating}.

Another approach in developing efficient deep learning methods is to redesign computationally expensive layers. For example, \citet{shift} replace standard convolution with a `Shift' layer that consumes zero flops during inference. \citet{shuffle} and 
\citet{shuffle2} achieve efficiency using a channel shuffle operation. A complementary solution for making deep neural networks suitable for embedded devices is to compress the model post training. This method doesn't require significant changes in architectural design. Notable examples of this approach include hashing \cite{hashing}, quantization \cite{quantization,quantization2} and factorization \cite{factor}. 

\section{MobiVSR: End-to-end Lip Reading with Fewer Parameters} \label{section:architecture}

\begin{figure}
\centering
 \includegraphics[width=0.90\linewidth]{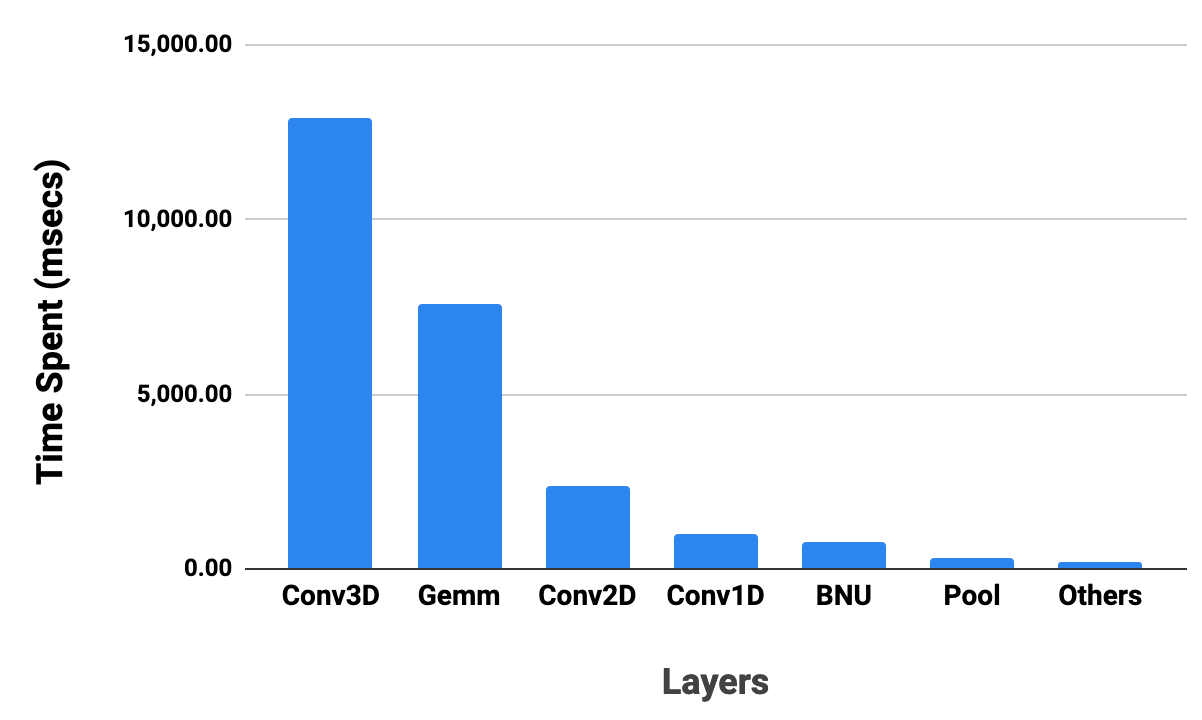}
 \caption{Time spent per layer of the lip reading architecture proposed in \cite{lipres}, profiled using cProfile.}
 \label{figure:profiling}
\end{figure}

This section describes the architecture of MobiVSR along with the explanations behind its design choices.

\begin{table*}[htbp]
\centering
\scalebox{0.8}{\begin{tabular}{@{}ccccccc@{}}
\toprule
\textbf{Model} & \textbf{\begin{tabular}[c]{@{}c@{}}Size \\ (in MB)\end{tabular}} & \textbf{\begin{tabular}[c]{@{}c@{}}Parameter count\\ (in millions)\end{tabular}} & \textbf{\begin{tabular}[c]{@{}c@{}}Memory\\ access\\ (in thousands)\end{tabular}} & \textbf{\begin{tabular}[c]{@{}c@{}}FLOPs\\ (in billions)\end{tabular}} & \textbf{\begin{tabular}[c]{@{}c@{}}Top-1\\ accuracy\end{tabular}} & \textbf{\begin{tabular}[c]{@{}c@{}}Top-3\\ accuracy\end{tabular}} \\ \midrule
\begin{tabular}[c]{@{}c@{}}LSTM + ResNet\\ (SOTA)\end{tabular} & 130.0 & 25.1 & 56.3 & 290 & 83 & 99.8 \\
LRW Baseline & 43.2 & 8.7 & 44.0 & 95.7 & 61.0 & 78.0 \\
\midrule
MobiVSR-1 & \textbf{17.8} & \textbf{4.5} & \textbf{35.3} & \textbf{11.0} & 72.2 & 88.0 \\
MobiVSR-2 & 20.8 & 5.2 & 37.3 & 20.1 & 73.0 & 89.0 \\
MobiVSR-3 & 23.6 & 5.9 & 38.9 & 29.5 & 73.4 & 90.2 \\
MobiVSR-4 & 26.5 & 6.6 & 40.4 & 40.1 & 74.0 & 91.0 \\
MobiVSR-10 & 43.9 & 10.8 & 51.5 & 92.4 & 77.1 & 96.1 \\
MobiVSR-11 & 46.8 & 11.5 & 53.3 & 99.8 & 77.5 & 97.3 \\
 \bottomrule
\end{tabular}}
\caption{\label{table:results} Comparison of accuracy, computational complexity, and memory footprint of MobiVSR (with different $\alpha$), with the LRW Baseline and the state-of-the-art model. \textbf{Note:} MobiVSR-1 denotes the model with $\alpha=1$}
\end{table*}

\subsection{The Problem}
\label{section:the problem}
%Here we briefly explain the problems before going into the experiments part. 
A brief overview of lipreading models is given in the Table \ref{table:results}. As shown in the table, the size of SOTA model is 130 MBs with a in-memory size of 56.3 MBs. It takes 290 billion FLOPs. With these parameters, it produces a top-3 accuracy of 99.8\%. LRW baseline presented in the paper \cite{lrw} is also given. However, as mentioned in Section \ref{section:intro}, the specifications of both the models are expensive from deployment perspectives, more so for mobile environments. Thus, after observing these and several other language recognition architectures \cite{kumar2018mylipper, kumar2018harnessing, petridis2016deep, petridis2017end}, we identified the following bottlenecks which limit their deployment avenues:

\noindent \textbf{$\bullet$} All of them use 3D convolutions for processing videos and it is the most compute intensive layer during inference. Figure~\ref{figure:profiling} shows the average percent of inference time spent per layer in the state-of-the-art lip reading system \cite{lipres}. Thus, this operation, in itself, increases their size and parameter count by several folds (as shown in Section \ref{section:experiments}).

\noindent \textbf{$\bullet$} Most of them use some form of RNNs but RNNs due to their non-parallelizable operations increase the inference time and imposes heavy costs of memory \cite{qrnn}. 

\noindent \textbf{$\bullet$} Generally, deep learning practitioners have the tendency to increase number of layers in the hope of getting better performance. This, was also a common observation with lipreading models. For example, the SOTA model contains 51 layers. However, parameter count and memory calls are directly correlated with the number of layers and play a major part in making the models non-usable for mobile environments.

\subsection{MobiVSR Architectural Design Choices}
The problems presented above (Section \ref{section:the problem}) are the common problems faced by all video recognition tasks. Thus, these problems become our motivation for the design choices of producing an efficient video recognition model, namely, MobiVSR.

\noindent \textbf{Challenge 1. Remove or optimize 3D convolutions}: 3D convolutions is a front-end technique in video processing tasks since it can combine information across both time and space \cite{lipres}. We observed, doing away with it deteriorates model accuracy. Therefore, optimizing 3D convolution becomes highly important. Inspired by \cite{mobilenet1}, where they converted 2D convolution into a sum of depthwise-separable and pointwise convolutions, we generalize 3D convolutions and use this to optimize the front end of the network in our architecture. 

Taking a cue from \cite{depthwise3d} and \cite{mobilenet1}, we optimize 3D and 2D convolutions with the perspective of making video recognition tasks more efficient. We do it by replacing naive convolutions with depthwise-separable convolutions. By doing this, we get a basic organized deep learning unit, which we call LipRes block. We show its applicability in a language recognition problems using videos (Table \ref{table:results}).

% Added after ACL reviews -NILAY
While using depthwise-separable convolutions in 3D, as the first step in any depthwise layer, an important factor to consider was the dimensions of pointwise convolution used. If we represent dimensions as $time \times height \times width$, then the issue is whether to use pointwise convolution of dimension $1 \times 1 \times 1$ (fully pointwise kernel) or $T \times 1 \times 1$ (partial pointwise kernel). The former has the advantage of reducing parameter count more than the latter. While designing the architecture of MobiVSR we found that the latter yields significantly better accuracy as compared to the fully pointwise kernel. We believe this is because in the partial pointwise convolution kernel, the time modality remains intact which is not the case in the fully pointwise kernel. Using $1 \times 1 \times 1$ kernel extracts information independently from each time frame rather than considering multiple frames. 

\noindent \textbf{Challenge 2. Avoid RNNs} : Recent research indicates that temporal convolutions can be used in place of RNNs without significant loss of accuracy \cite{cnnoverrnn}. Temporal convolutions offer several advantages. They increase parallelism since convolution can process multiple time-steps at once. They also have flexible receptive fields \cite{mac2018locally}, and can control model memory usage. Therefore we use 1-D temporal convolution in place of a RNN for modeling temporal features.

\noindent \textbf{Challenge 3. Reduce Parameter Count in Convolution Filters} \label{strategy:convolution} : First, since the kernel parameter count increases quadratically with the kernel size, we had to use small convolution filter sizes of $3 \times 3$ in MobiVSR. Due to this constraint, we gained a bonus boost thanks to the modern deep learning frameworks which use several algorithms that optimize the number of operations required for a convolution operation \cite{convocomplexity} with a small filter size \cite{fft1,fft2,cudnn,winograd}. Second, we use the depthwise-separable convolutions \cite{mobilenet1}, for reducing the time and computational complexity of the model. 

% \noindent \textbf{Our Strategy}
\begin{figure}[htbp]
\includegraphics[width=7cm]{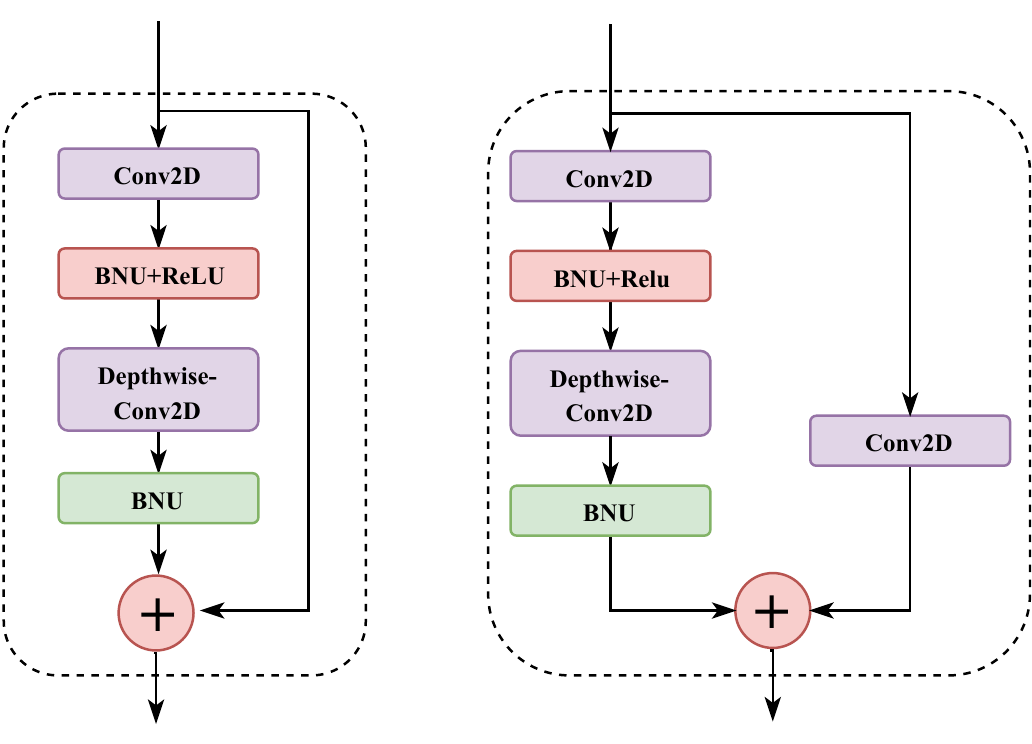}
\centering
\caption{LipRes Blocks: (1) The first LipRes block is for keeping the size of the input constant. It has been used in the first subgraph in MobiVSR. (2) The second LipRes block is for halving the size of the input. It is used in the second, third and the fourth subgraphs in MobiVSR. Thus, increasing alpha by 1, increases the first LipRes block by one and the second LipRes block by 3.}
\label{fig:lipRes}
\end{figure}

\noindent \textbf{Use Residual Connections} \label{strategy:residual}: Since increasing the depth of a deep network to increase accuracy adds additional computational complexity and more memory calls, we use residual connections (ResNet blocks) as suggested in \cite{resnet}. These connections are used extensively inside the LipRes block as shown in Fig. \ref{fig:lipRes}. LipRes has a residual structure similar to the ResNet blocks. Each block consists of depthwise-separable convolutions and ReLU \cite{relu} non linearity, and a parallel skip connection. The use of depthwise separable convolutions and residual connections helps to reduce the parameter count and cut memory calls. The skip connection also has a convolution with stride two whenever the output is supposed to be spatially down-sampled.

Additionally, we use batch normalization \cite{batchnorm} for regularization. However, contrary to the common paradigm of using batch normalization after every convolution kernel, we use batch normalization scantily as the smaller size of the network has a regularizing effect during training.

\noindent \textbf{Challenge 4. Introduction of LipRes block and varying $\alpha$}: 
We incorporate solutions to the challenges 1, 2 and 3 in the LipRes block. Through experimentation, we observed that increasing the number of LipRes blocks increased the accuracy but at a cost of making the model heavier and more energy intensive. Thus, we realized that number of LipRes blocks could be leveraged to get a balance between the accuracy and the model size. In the experiments, we show the use of LipRes blocks and also vary its number in order to demonstrate it as a performance-size trade-off handle. For the latter part, we represent the number of blocks by a hyperparameter $\alpha$ which is varied.

\subsection{MobiVSR Architecture}
\label{subsec:macroarch}
\begin{figure*}[htbp]
 \includegraphics[width=\textwidth]{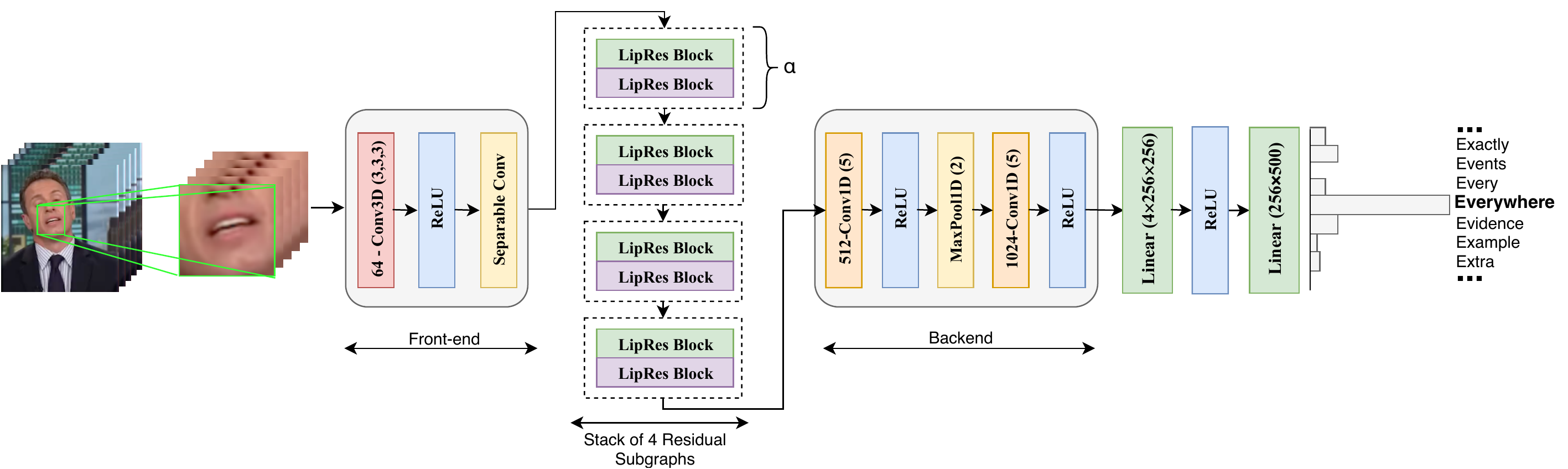}
 \caption{MobiVSR architecture}
 \label{fig:MobiVSR}
\end{figure*}
% AS - in general, in this section the goal should be to explain *why* choices were made and/or *what* the parts of the architecture achieve. This will help the NLP audience. For example, what parts do low level visual feature extraction? What parts identify visimes? What parts (if any) are the equivalent of language modeling in a classic ASR architecture?
The MobiVSR architecture is shown in Figure \ref{fig:MobiVSR}.
The MobiVSR model essentially maps visemes (basic units of visual speech) to textual units (\textit{i.e.}, characters/words). Thus, with this in mind, it can be divided into 3 broad parts: (1) a frontend three dimensional convolution part whose function is to extract low level features from visemes; (2) a middle stack of variable sized residual subgraphs whose function is to use those low level features to infer high level features; and (3) a backend consisting of temporal convolutions which essentially functions as a language model of a classical automatic speech recognition system (ASR) by integrating the high-level features to get text out of visemes. Finally, there are two fully connected neural network layers that outputs class probabilities, thus converting abstract grapheme predictions to their probabilities.

The front-end part of MobiVSR consists of two 3D depthwise separable convolution layers. We use two depthwise separable layers with kernel size ${\scriptstyle 3 \times 3 \times 3}$. Each of these layers downsize the input along spatial dimension by two; we implement this layer by using ${\scriptstyle 3 \times 3 \times 3}$ group convolution, with the number of groups being equal to the number of input channels followed by a point-wise spatial convolution kernel of ${\scriptstyle 3 \times 1 \times 1}$. 

The middle stack of MobiVSR is subdivided into four subgraphs. Each subgraph has $\alpha$ LipRes blocks. Here $\alpha$ is a hyperparameter which one can vary to change the depth of the model. The results pertaining to these are shown in the Table \ref{table:results}, where increasing $\alpha$ increases accuracy but at the cost of using more parameters and vice-versa. Thus, the handle of $\alpha$ makes MobiVSR architecture flexible enough for different applications and environments.

The backend of the model is used to integrate the features across time and provide the word probabilities. Thus, it uses two temporal convolution layers with a Maxpool downsampling sandwiched in between. For performance reasons as outlined above we do not use RNN layers, in contrast to \cite{lipres}. Finally MobiVSR uses fully connected layers with softmax activation to generate class probabilities for the 500 words in the LRW dataset.

\section{Experimental Setup and Results}
\label{experiments}

\subsection{Data}
\label{data}

We base our experiments on the large publicly available speaker-independent Lip Reading in the Wild (LRW) database \cite{lrw}. The LRW database contains 1000 utterances each for a collection of 500 different words in the training set. For testing and validation, the dataset has 50 video clips per class. Each video is challenging because of the high variance of head pose and illumination; therefore, in addition to being one of the few VSR data sets of size, LRW is a good proxy for mobile lip reading data. The video clips are of 29 frames (1.16 seconds) in length, and the speaker mouth region of interest (ROI) is placed at the center of each frame. 

A set of 29 consecutive frames (256 $\times$ 256 pixels) is sampled from each video of the LRW Dataset. We then extract the mouth ROI from these RGB frames. As the LRW Dataset is face centered, we achieve this step by cropping a $96\times96$ pixel window image segment from the center of each frame. Finally, the cropped frame segments are converted to gray scale and stored as numpy matrices \cite{numpy}. This numpy matrix is then fed to all networks as input. 

\subsection{Experiments}
\label{section:experiments}

We train MobiVSR on a NVIDIA Titan X GPU for 50 epochs using six different settings for $\alpha$ (1, 2, 3, 4, 10 and 11). The results are summarized in Table \ref{table:results}. % Fig. \ref{fig:internal} shows a accuracy vs. parameter graph for MobiVSR models with different $\alpha$. %To increase accuracy, one can increase the number of LipRes blocks by increasing the value of $\alpha$. On the other hand one can get a smaller and more efficient model at the cost of some accuracy by reducing $\alpha$.
We compare MobiVSR with other word-level lip reading models on saved model size, number of parameters, memory required during inference and number of floating point operations (FLOPs). To ensure consistency in comparing model sizes, we converted each model to ONNX format\footnote{https://onnx.ai}. We calculate inference speeds on an Intel i3 processor
%\footnote{i3 is not a mobile-first processor but it is also not a GPU environment. We {\bf have} successfully run the MobiVSR model on a mobile device and can demonstrate this at the conference.} 
and average over 5000 runs. The calculations of memory access and number of floating point operations in different layers are described in Appendix and summarized in Table \ref{table:memaccflop}. We ignore the effects of applying non-linearities, batch-normalization and bias terms in these calculations as they do not make significant contributions in comparison to matrix multiplications and convolutions. Interested readers can check the appendix for an example and further details regarding the derivation of the expressions as given in Table \ref{table:memaccflop}. Energy consumption of various models given in  Table \ref{table:energyconsumed} has been estimated by multiplying the energy values of various multiplication, addition and memory access provided in Table \ref{table:energy} with their corresponding numbers in Table \ref{table:memaccflop}. Furthermore to obtain $CO_2$ emission estimates, we multiplied the energy consumption values with the average $CO_2$ emission per kWh of energy as suggested in \cite{emission}.

\begin{table*}[htbp]
\centering
\scalebox{0.8}{\begin{tabular}{@{}lll@{}}
\toprule
\multicolumn{1}{c}{Layer} & \multicolumn{1}{c}{Memory Access} & \multicolumn{1}{c}{\begin{tabular}[c]{@{}c@{}}Floating point operations\\ (FLOPs)\end{tabular}} \\ \midrule
Conv2D & $K^2C_iC_o+V_i\cdot(K^2Co) +V_o$ & $2(K^2C_i) \cdot V_o$ \\
Conv3D & $K^2TC_iC_o+ V_i\cdot(K^2C_o) \cdot T+V_o$ & $2(K^2TC_i)\cdot V\_o$ \\
\begin{tabular}[c]{@{}l@{}}Depthwise Separable\\ Conv2D\end{tabular} & $C_i \cdot(K^2 + C_o)+V_i\cdot(K^2+C_o)+V_o$ & $2C_i \cdot(\frac{K^2}{C_o}+1)\cdot V_o$ \\
\begin{tabular}[c]{@{}l@{}}Depthwise Separable\\ Conv3D\end{tabular} & $C_i \cdot(K^2+ C_o) \cdot T+V_i \cdot(K^2 + C_o)\cdot T+V_o$ & $2C_i\cdot(\frac{K^2}{C_o}+1)\cdot T\cdot V_o$ \\
\begin{tabular}[c]{@{}l@{}}Fully Connected\end{tabular} &$IQ+V_i+V_o$ &$2IQ$ \\
\bottomrule
\end{tabular}}
\caption{\label{table:memaccflop}Number of memory access and FLOPs associated with different layers. $V_i$ and $V_o$ are the input and output volume respectively. $C_i$ and $C_o$ are the input and output channel dimensions. $K\times K$ is the 2D convolution kernel while $K \times K \times T$ is the 3D convolution kernel}
\end{table*}

\begin{table}[]
\centering
\scalebox{0.8}{
\begin{tabular}{@{}lcc@{}}
\toprule
\multicolumn{1}{c}{\textbf{Model}} & \textbf{\begin{tabular}[c]{@{}c@{}}Energy Consumed\\ per Inference (milli Joules)\end{tabular}} & \textbf{\begin{tabular}[c]{@{}c@{}}$CO_{2}$ emitted\\ per Inference\\  (mg)\end{tabular}} \\ \midrule
MobiVSR-1                          & 25.37                                                                                           & 3.21                                                                                       \\
MobiVSR-2                          & 46.30                                                                                           & 5.85                                                                                       \\
MobiVSR-3                          & 67.92                                                                                           & 8.59                                                                                       \\
MobiVSR-4                          & 92.31                                                                                           & 11.67                                                                                      \\
MobiVSR-10                         & 212.62                                                                                          & 26.89                                                                                      \\
MobiVSR-11                         & 229.64                                                                                          & 29.01                                                                                      \\
LSTM + ResNet                      & 667.11                                                                                          & 84.38                                                                                      \\
LRW Baseline                       & 229.39                                                                                          & 29.01                                                                                      \\ \bottomrule
\end{tabular}
}
\caption{\label{table:energyconsumed} Comparison of models on the basis of energy consumed and $CO_{2}$ emission per inference.}
\end{table}

% AS - I suggest to reduce the size of this and the next section (computing FLOPs) so as to leave more room for an error analysis of where the different models fall over on this data set.

\section{Analysis of Architecture and Results}
In this section, we explain the contributions made towards various architectural decisions in results. We also present qualitative evaluation of the results for visual recognition of language.

% CO2 EMISSION data
% JOULES calculated using table \ref{table:energy}
% CO2 emission calculating method : avg emission / joule obtained from \cite{emission} 

\subsection{Quantitative Formulation}
\label{section:quant}
\begin{figure}
\includegraphics[width=7cm]{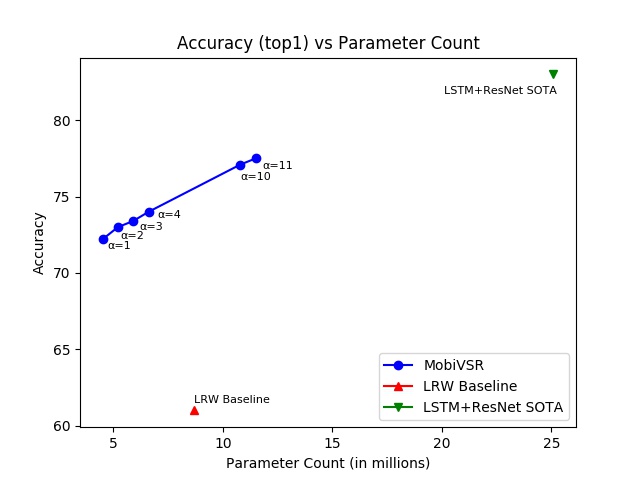}
\centering
\caption{Accuracy vs Parameter Count plot for various values of $\alpha$}
\setlength{\belowcaptionskip}{-15pt}
\label{fig:accuracyVsParam}
\end{figure}

From the Table \ref{table:results}, it can be inferred that MobiVSR-1 achieves 72.2\% accuracy on LRW dataset and reaches to an \textit{uncompressed size} of 17.8 MB. As expected, the accuracy increases by increasing the number of LipRes blocks (Section \ref{section:architecture}). However, several interesting trends can be inferred from the results. One such is the accuracy to size ratio comparison. MobiVSR-1 has an accuracy to size ratio of 4.06 while the SOTA's value is 0.64. Thus, MobiVSR gives much higher accuracy per megabyte space as compared to any other model. Further, in case of MobiVSR the slope of increase of accuracy \textit{vs} model size is almost 3 MBs per 100 basis point increase in performance (Figure \ref{fig:accuracyVsParam}). The statistics of parameter count, memory access and FLOPs also follow the same trends. The values are (SOTA/MobiVSR-1) - FLOPs (0.29/6.56), memory access (1.47/2.04) and parameter count (3.31/16.04). Most of these values show improvements of the order of 10x.

%ACCURACY vs ALPHA figure
 
\begin{table}[]
\centering
\caption{Some sample words and what are they getting confused against. The first two rows contain maximally confusing words, the next two contain average cases and the last two the least confused words.}% on the OuluVS2 database for the first 25 speakers}
\label{table:confusion_explained}
\scalebox{0.85}{\begin{tabular}{ll} \hline
\textbf{Word}  & \textbf{Confusion} \\ \hline
Benefits & Benefit\\
Price & Press\\ \hline
Words	& Workers \\
Action & Actually\\ \hline
About & Afternoon \\
About & Temperatures \\ \hline
\end{tabular}
\begin{tabular}{ll}\hline
\textbf{Word}  & \textbf{Confusion} \\ \hline  
Spent & Spend\\
Living & Giving\\ \hline
These & Years \\
Community & Abuse \\ \hline
About & Weapons \\ 
About & Westminster \\ \hline
\end{tabular}
}\end{table}

\begin{figure}[htbp]
\includegraphics[width=7cm]{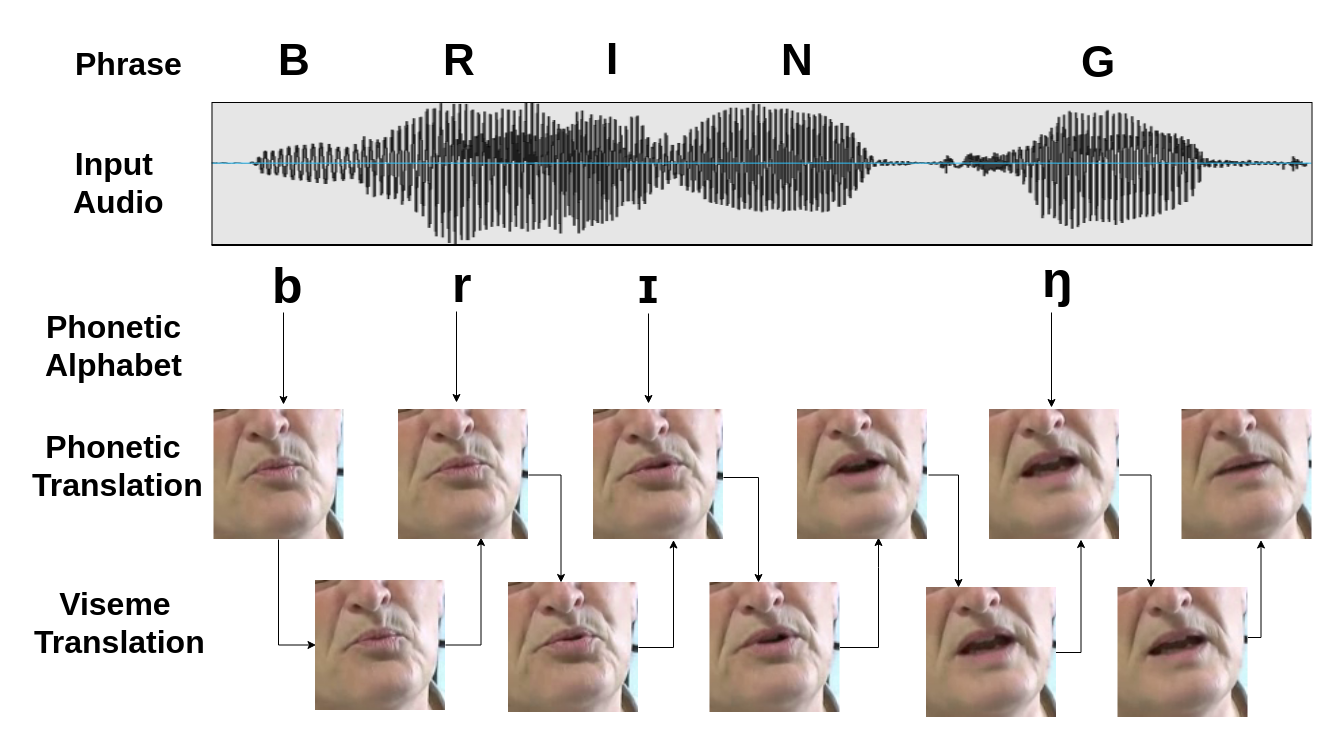}
\centering
 \setlength{\belowcaptionskip}{-15pt}
\caption{Visemic-phonemic correspondence for the word ``Bring''}
\label{fig:bring_img}
\end{figure}

\subsection{Qualitative Formulation}
While it is quintessential for the model to focus on parameters like memory access, size, parameter count, it is also essential for us to analyse the strengths and weaknesses of the model. Hence we analysed the performance of the model by looking at some specific cases. 

\begin{figure}[htbp]
\includegraphics[width=7cm]{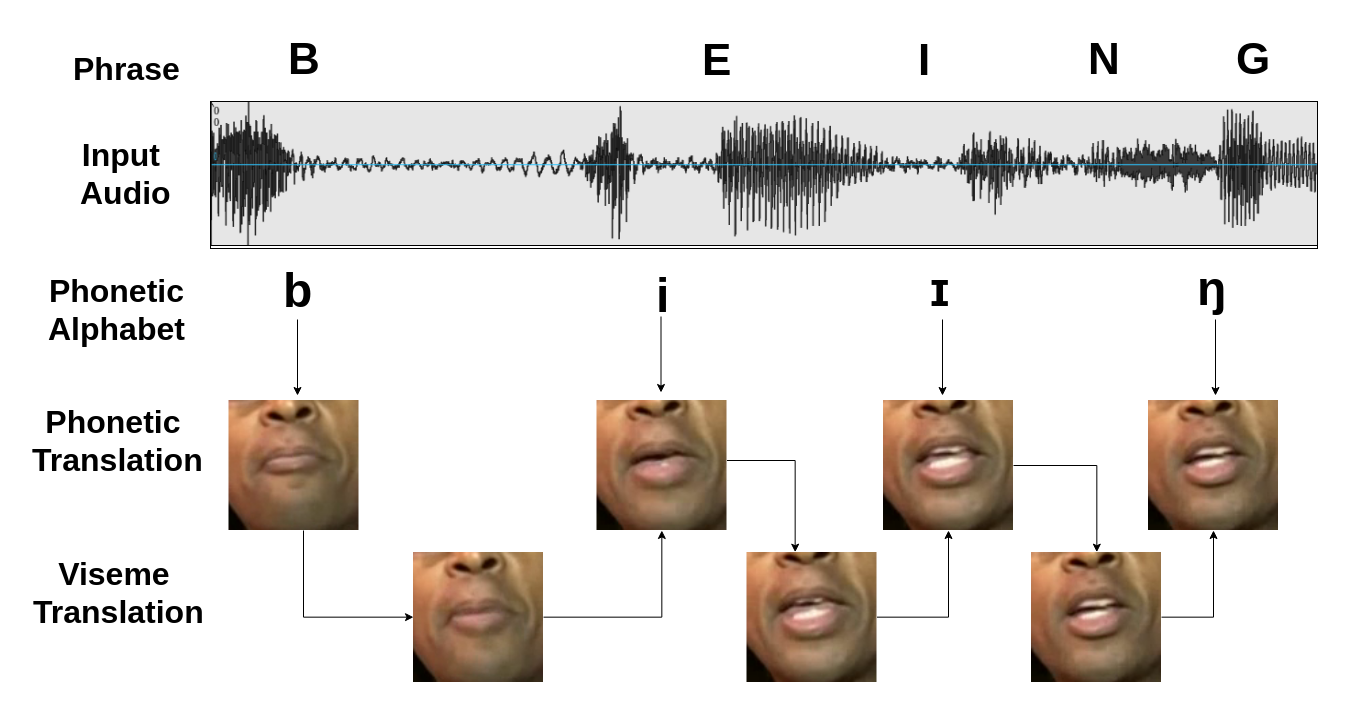}
\centering
 \setlength{\belowcaptionskip}{-15pt}
\caption{Visemic-phonemic correspondence for the word ``Being''}
\label{fig:being_img}
\end{figure}

While the percentage of true positive predicted by the model is high, there are a few interesting failure cases as well. We observed that certain samples were mo in the dataset , in the dataset samples and human evaluation experiments, some words were being confused with other words. We found out that these are those words which have some common visemes with other words. For example, take the case of these two words: \textit{`bring'} and \textit{`being'} as presented in Figures \ref{fig:bring_img} and \ref{fig:being_img}. `Bring' has the following visemes: \{\textit{E,A,V4,H}\} and `Being' has the following: \{\textit{E,V4,V4,H}\} \cite{neti2000audio}. As can be seen in the figures, three out of four visemes are common in both of them. The fourth ones which are different are spoken from within the mouth, hence are difficult to capture using a camera. The model confuses between them 80\% of the time. We observed similar behaviour with these pairs as well: \{\textit{Billions,Millions}\}, \{\textit{Having, Heavy}\}, \{\textit{General, Several}\}, \textit{etc}. 

Several such cases are documented in the Table \ref{table:confusion_explained}. The last two rows in the table contain those words which are not confused with any other words. As can be observed (by speaking those words) that those words do not contain many phonemes or even visemes in common. This should be the reason why they aren't confused. Using these error cases, we discover that model, when it fails, fails due to the signals which cannot be captured using a camera only. For example, the sound of \textit{`ba'} in billions and \textit{`ma'} in millions belong to the same viseme class but map to different phonemes. Thus, a camera alone cannot capture and differentiate between these two, however, when coupled with an audio-recording device, signals from both of them combined can potentially help to discriminate amongst them \footnote{Since LRW is a dataset which is quite close to real environment, we perform some experiments on some people who were not present in the dataset. Images of one such sample is given in the appendix.}.

\section{Conclusion and Future Work}
In this paper we introduced MobiVSR, a deep neural network model designed to perform word level visual speech recognition in resource constrained devices. We showed how MobiVSR uses $6 \times$ less parameters than the state-of-the-art model and can be compressed to 6MB after quantization. Moreover it can be modified using a tuneable hyperparameter to balance accuracy and efficiency for different use cases. As mentioned earlier, mobile-centric lip reading systems have enormous utility in the society. We hope that this paper inspires other researchers to create similar and even more efficient models considering the social impact such applications can have. 

%In the future we would like to use our system in order to develop assistive technologies for people having speech problems such as patients suffering from Dysarthria. We would also like to develop a pose-invariant model for hand held devices by extending our current work and also test our current model in other datasets. Further, we would like to perform experiments to study how we can use our model in noisy environments and real-life scenarios such as improving speech recognition during driving conditions and video conferencing.

\bibliography{mobivsr}

\begin{thebibliography}{62}
\expandafter\ifx\csname natexlab\endcsname\relax\def\natexlab#1{#1}\fi

\bibitem[{Assael et~al.(2016)Assael, Shillingford, Whiteson, and
  De~Freitas}]{assael2016lipnet}
Yannis~M Assael, Brendan Shillingford, Shimon Whiteson, and Nando De~Freitas.
  2016.
\newblock Lipnet: End-to-end sentence-level lipreading.
\newblock \emph{arXiv preprint arXiv:1611.01599}.

\bibitem[{Bai et~al.(2018)Bai, Kolter, and Koltun}]{cnnoverrnn}
Shaojie Bai, J~Zico Kolter, and Vladlen Koltun. 2018.
\newblock An empirical evaluation of generic convolutional and recurrent
  networks for sequence modeling.
\newblock \emph{arXiv preprint arXiv:1803.01271}.

\bibitem[{Bernstein et~al.(1998)Bernstein, Demorest, and
  Tucker}]{bernstein1998makes}
Lynne~E Bernstein, Marilyn~E Demorest, and Paula~E Tucker. 1998.
\newblock What makes a good speechreader? first you have to find one.
\newblock \emph{Hearing by eye II: Advances in the psychology of speechreading
  and auditory-visual speech}, pages 211--227.

\bibitem[{Bradbury et~al.(2016)Bradbury, Merity, Xiong, and Socher}]{qrnn}
James Bradbury, Stephen Merity, Caiming Xiong, and Richard Socher. 2016.
\newblock Quasi-recurrent neural networks.
\newblock \emph{arXiv preprint arXiv:1611.01576}.

\bibitem[{Bulwer(1648)}]{deafanddumb}
John Bulwer. 1648.
\newblock \href
  {http://www.acsu.buffalo.edu/~duchan/new_history/early_modern/bulwer.html}
  {Philocopus, or the deaf and dumbe mans friend , london: Humphrey and
  moseley.}

\bibitem[{Carbonfund.org(2019)}]{emission}
Carbonfund.org. 2019.
\newblock \href {https://carbonfund.org/how-we-calculate/} {Co2 emission
  calculation}.

\bibitem[{Chellapilla et~al.(2006)Chellapilla, Puri, and
  Simard}]{convocomplexity}
Kumar Chellapilla, Sidd Puri, and Patrice Simard. 2006.
\newblock High performance convolutional neural networks for document
  processing.
\newblock In \emph{Tenth International Workshop on Frontiers in Handwriting
  Recognition}. Suvisoft.

\bibitem[{Chen and Rao(1998)}]{chen1998audio}
Tsuhan Chen and Ram~R Rao. 1998.
\newblock Audio-visual integration in multimodal communication.
\newblock \emph{Proceedings of the IEEE}, 86(5):837--852.

\bibitem[{Chen et~al.(2015)Chen, Wilson, Tyree, Weinberger, and Chen}]{hashing}
Wenlin Chen, James Wilson, Stephen Tyree, Kilian Weinberger, and Yixin Chen.
  2015.
\newblock Compressing neural networks with the hashing trick.
\newblock In \emph{International Conference on Machine Learning}, pages
  2285--2294.

\bibitem[{Chetlur et~al.(2014)Chetlur, Woolley, Vandermersch, Cohen, Tran,
  Catanzaro, and Shelhamer}]{cudnn}
Sharan Chetlur, Cliff Woolley, Philippe Vandermersch, Jonathan Cohen, John
  Tran, Bryan Catanzaro, and Evan Shelhamer. 2014.
\newblock cudnn: Efficient primitives for deep learning.
\newblock \emph{arXiv preprint arXiv:1410.0759}.

\bibitem[{Chung et~al.(2017)Chung, Senior, Vinyals, and
  Zisserman}]{chung2017lip}
Joon~Son Chung, Andrew Senior, Oriol Vinyals, and Andrew Zisserman. 2017.
\newblock Lip reading sentences in the wild.
\newblock In \emph{2017 IEEE Conference on Computer Vision and Pattern
  Recognition (CVPR)}, pages 3444--3453. IEEE.

\bibitem[{Chung and Zisserman(2016{\natexlab{a}})}]{chung2016lip}
Joon~Son Chung and Andrew Zisserman. 2016{\natexlab{a}}.
\newblock Lip reading in the wild.
\newblock In \emph{Asian Conference on Computer Vision}, pages 87--103.
  Springer.

\bibitem[{Chung and Zisserman(2016{\natexlab{b}})}]{lrw}
Joon~Son Chung and Andrew Zisserman. 2016{\natexlab{b}}.
\newblock Lip reading in the wild.
\newblock In \emph{Asian Conference on Computer Vision}, pages 87--103.
  Springer.

\bibitem[{Cornu and Milner(2015)}]{cornu2015reconstructing}
Thomas~Le Cornu and Ben Milner. 2015.
\newblock Reconstructing intelligible audio speech from visual speech features.
\newblock In \emph{Sixteenth Annual Conference of the International Speech
  Communication Association}.

\bibitem[{Demorest and Bernstein(1991)}]{demorest1991computational}
Marilyn~E Demorest and Lynne~E Bernstein. 1991.
\newblock Computational explorations of speechreading.
\newblock \emph{Journal of the Academy of Rehabilitative Audiology}.

\bibitem[{Dodd and Campbell(1987)}]{dodd1987hearing}
Barbara~Ed Dodd and Ruth~Ed Campbell. 1987.
\newblock \emph{Hearing by eye: The psychology of lip-reading.}
\newblock Lawrence Erlbaum Associates, Inc.

\bibitem[{Ephrat et~al.(2017)Ephrat, Halperin, and Peleg}]{ephrat2017improved}
Ariel Ephrat, Tavi Halperin, and Shmuel Peleg. 2017.
\newblock Improved speech reconstruction from silent video.
\newblock In \emph{Proceedings of the IEEE International Conference on Computer
  Vision}, pages 455--462.

\bibitem[{Han et~al.(2015)Han, Mao, and Dally}]{modelcompression}
Song Han, Huizi Mao, and William~J Dally. 2015.
\newblock Deep compression: Compressing deep neural networks with pruning,
  trained quantization and huffman coding.
\newblock \emph{arXiv preprint arXiv:1510.00149}.

\bibitem[{He et~al.(2016)He, Zhang, Ren, and Sun}]{resnet}
Kaiming He, Xiangyu Zhang, Shaoqing Ren, and Jian Sun. 2016.
\newblock Deep residual learning for image recognition.
\newblock In \emph{Proceedings of the IEEE conference on computer vision and
  pattern recognition}, pages 770--778.

\bibitem[{Horowitz(2014)}]{energystat}
Mark Horowitz. 2014.
\newblock 1.1 computing's energy problem (and what we can do about it).
\newblock In \emph{Solid-State Circuits Conference Digest of Technical Papers
  (ISSCC), 2014 IEEE International}, pages 10--14. IEEE.

\bibitem[{Howard et~al.(2017)Howard, Zhu, Chen, Kalenichenko, Wang, Weyand,
  Andreetto, and Adam}]{mobilenet1}
Andrew~G Howard, Menglong Zhu, Bo~Chen, Dmitry Kalenichenko, Weijun Wang,
  Tobias Weyand, Marco Andreetto, and Hartwig Adam. 2017.
\newblock Mobilenets: Efficient convolutional neural networks for mobile vision
  applications.
\newblock \emph{arXiv preprint arXiv:1704.04861}.

\bibitem[{Huggins-Daines et~al.(2006)Huggins-Daines, Kumar, Chan, Black,
  Ravishankar, and Rudnicky}]{huggins2006pocketsphinx}
David Huggins-Daines, Mohit Kumar, Arthur Chan, Alan~W Black, Mosur
  Ravishankar, and Alexander~I Rudnicky. 2006.
\newblock Pocketsphinx: A free, real-time continuous speech recognition system
  for hand-held devices.
\newblock In \emph{2006 IEEE International Conference on Acoustics Speech and
  Signal Processing Proceedings}, volume~1, pages I--I. IEEE.

\bibitem[{Iandola et~al.(2016)Iandola, Han, Moskewicz, Ashraf, Dally, and
  Keutzer}]{squeezenet}
Forrest~N Iandola, Song Han, Matthew~W Moskewicz, Khalid Ashraf, William~J
  Dally, and Kurt Keutzer. 2016.
\newblock Squeezenet: Alexnet-level accuracy with 50x fewer parameters and< 0.5
  mb model size.
\newblock \emph{arXiv preprint arXiv:1602.07360}.

\bibitem[{Ioffe and Szegedy(2015)}]{batchnorm}
Sergey Ioffe and Christian Szegedy. 2015.
\newblock Batch normalization: Accelerating deep network training by reducing
  internal covariate shift.
\newblock \emph{arXiv preprint arXiv:1502.03167}.

\bibitem[{Krizhevsky et~al.(2012)Krizhevsky, Sutskever, and Hinton}]{alexnet}
Alex Krizhevsky, Ilya Sutskever, and Geoffrey~E Hinton. 2012.
\newblock Imagenet classification with deep convolutional neural networks.
\newblock In \emph{Advances in neural information processing systems}, pages
  1097--1105.

\bibitem[{Kumar et~al.(2018{\natexlab{a}})Kumar, Aggarwal, Nawal, Satoh, Shah,
  and Zimmermann}]{kumar2018harnessing}
Yaman Kumar, Mayank Aggarwal, Pratham Nawal, Shin'ichi Satoh, Rajiv~Ratn Shah,
  and Roger Zimmermann. 2018{\natexlab{a}}.
\newblock Harnessing ai for speech reconstruction using multi-view silent video
  feed.
\newblock In \emph{2018 ACM Multimedia Conference on Multimedia Conference},
  pages 1976--1983. ACM.

\bibitem[{Kumar et~al.(2018{\natexlab{b}})Kumar, Jain, Salik, ratn Shah,
  Zimmermann, and Yin}]{kumar2018mylipper}
Yaman Kumar, Rohit Jain, Mohd Salik, Rajiv ratn Shah, Roger Zimmermann, and
  Yifang Yin. 2018{\natexlab{b}}.
\newblock Mylipper: A personalized system for speech reconstruction using
  multi-view visual feeds.
\newblock In \emph{2018 IEEE International Symposium on Multimedia (ISM)},
  pages 159--166. IEEE.

\bibitem[{Lavin and Gray(2016)}]{winograd}
Andrew Lavin and Scott Gray. 2016.
\newblock Fast algorithms for convolutional neural networks.
\newblock In \emph{Proceedings of the IEEE Conference on Computer Vision and
  Pattern Recognition}, pages 4013--4021.

\bibitem[{Lebedev et~al.(2014)Lebedev, Ganin, Rakhuba, Oseledets, and
  Lempitsky}]{factor}
Vadim Lebedev, Yaroslav Ganin, Maksim Rakhuba, Ivan Oseledets, and Victor
  Lempitsky. 2014.
\newblock Speeding-up convolutional neural networks using fine-tuned
  cp-decomposition.
\newblock \emph{arXiv preprint arXiv:1412.6553}.

\bibitem[{Ma et~al.(2018)Ma, Zhang, Zheng, and Sun}]{shuffle2}
Ningning Ma, Xiangyu Zhang, Hai-Tao Zheng, and Jian Sun. 2018.
\newblock Shufflenet v2: Practical guidelines for efficient cnn architecture
  design.
\newblock In \emph{Proceedings of the European Conference on Computer Vision
  (ECCV)}, pages 116--131.

\bibitem[{Mac et~al.(2018)Mac, Joshi, Yeh, Xiong, Feris, and
  Do}]{mac2018locally}
Khoi-Nguyen~C Mac, Dhiraj Joshi, Raymond~A Yeh, Jinjun Xiong, Rogerio~R Feris,
  and Minh~N Do. 2018.
\newblock Locally-consistent deformable convolution networks for fine-grained
  action detection.
\newblock \emph{arXiv preprint arXiv:1811.08815}.

\bibitem[{Marschark et~al.(1998)Marschark, LePoutre, and
  Bement}]{marschark1998mouth}
Marc Marschark, Dominique LePoutre, and Linda Bement. 1998.
\newblock \emph{Mouth movement and signed communication}.
\newblock Hove, United Kingdom: Psychology Press Ltd. Publishers.

\bibitem[{Mathieu et~al.(2013)Mathieu, Henaff, and LeCun}]{fft1}
Michael Mathieu, Mikael Henaff, and Yann LeCun. 2013.
\newblock Fast training of convolutional networks through ffts.
\newblock \emph{arXiv preprint arXiv:1312.5851}.

\bibitem[{Mori et~al.(2018)Mori, Tjandra, Sakti, and
  Nakamura}]{mori2018compressing}
Takuma Mori, Andros Tjandra, Sakriani Sakti, and Satoshi Nakamura. 2018.
\newblock Compressing end-to-end asr networks by tensor-train decomposition.
\newblock \emph{Proc. Interspeech 2018}, pages 806--810.

\bibitem[{Morse et~al.(1978)Morse, Pohlman, and Ravenel}]{intel8086}
Stephen~P Morse, William~B Pohlman, and Bruce~W Ravenel. 1978.
\newblock The intel 8086 microprocessor: a 16-bit evolution of the 8080.
\newblock \emph{Computer}, (6):18--27.

\bibitem[{Nair and Hinton(2010)}]{relu}
Vinod Nair and Geoffrey~E Hinton. 2010.
\newblock Rectified linear units improve restricted boltzmann machines.
\newblock In \emph{Proceedings of the 27th international conference on machine
  learning (ICML-10)}, pages 807--814.

\bibitem[{Neti et~al.(2000)Neti, Potamianos, Luettin, Matthews, Glotin,
  Vergyri, Sison, Mashari, and Zhou}]{neti2000audio}
Chalapathy Neti, Gerasimos Potamianos, Juergen Luettin, Iain Matthews, Herve
  Glotin, Dimitra Vergyri, June Sison, Azad Mashari, and Jie Zhou. 2000.
\newblock Audio-visual speech recognition.
\newblock In \emph{Final workshop}, pages 40--41.

\bibitem[{Ngiam et~al.(2011)Ngiam, Khosla, Kim, Nam, Lee, and
  Ng}]{ngiam2011multimodal}
Jiquan Ngiam, Aditya Khosla, Mingyu Kim, Juhan Nam, Honglak Lee, and Andrew~Y
  Ng. 2011.
\newblock Multimodal deep learning.
\newblock In \emph{Proceedings of the 28th international conference on machine
  learning (ICML-11)}, pages 689--696.

\bibitem[{Noda et~al.(2015)Noda, Yamaguchi, Nakadai, Okuno, and
  Ogata}]{noda2015audio}
Kuniaki Noda, Yuki Yamaguchi, Kazuhiro Nakadai, Hiroshi~G Okuno, and Tetsuya
  Ogata. 2015.
\newblock Audio-visual speech recognition using deep learning.
\newblock \emph{Applied Intelligence}, 42(4):722--737.

\bibitem[{Pachoud et~al.(2008)Pachoud, Gong, and Cavallaro}]{pachoud2008macro}
Samuel Pachoud, Shaogang Gong, and Andrea Cavallaro. 2008.
\newblock Macro-cuboid based probabilistic matching for lip-reading digits.
\newblock In \emph{2008 IEEE Conference on Computer Vision and Pattern
  Recognition}, pages 1--8. IEEE.

\bibitem[{Park et~al.(2018)Park, Boo, Choi, Shin, and Sung}]{park2018fully}
Jinhwan Park, Yoonho Boo, Iksoo Choi, Sungho Shin, and Wonyong Sung. 2018.
\newblock Fully neural network based speech recognition on mobile and embedded
  devices.
\newblock In \emph{Advances in Neural Information Processing Systems}, pages
  10642--10653.

\bibitem[{Petridis and Pantic(2016)}]{petridis2016deep}
Stavros Petridis and Maja Pantic. 2016.
\newblock Deep complementary bottleneck features for visual speech recognition.
\newblock In \emph{2016 IEEE International Conference on Acoustics, Speech and
  Signal Processing (ICASSP)}, pages 2304--2308. IEEE.

\bibitem[{Petridis et~al.(2017)Petridis, Wang, Li, and
  Pantic}]{petridis2017end}
Stavros Petridis, Yujiang Wang, Zuwei Li, and Maja Pantic. 2017.
\newblock End-to-end multi-view lipreading.
\newblock \emph{arXiv preprint arXiv:1709.00443}.

\bibitem[{Polino et~al.(2018)Polino, Pascanu, and Alistarh}]{quantization}
Antonio Polino, Razvan Pascanu, and Dan Alistarh. 2018.
\newblock Model compression via distillation and quantization.
\newblock \emph{arXiv preprint arXiv:1802.05668}.

\bibitem[{Prabhavalkar et~al.(2016)Prabhavalkar, Alsharif, Bruguier, and
  McGraw}]{prabhavalkar2016compression}
Rohit Prabhavalkar, Ouais Alsharif, Antoine Bruguier, and Lan McGraw. 2016.
\newblock On the compression of recurrent neural networks with an application
  to lvcsr acoustic modeling for embedded speech recognition.
\newblock In \emph{2016 IEEE International Conference on Acoustics, Speech and
  Signal Processing (ICASSP)}, pages 5970--5974. IEEE.

\bibitem[{Rastegari et~al.(2016)Rastegari, Ordonez, Redmon, and
  Farhadi}]{xnornet}
Mohammad Rastegari, Vicente Ordonez, Joseph Redmon, and Ali Farhadi. 2016.
\newblock Xnor-net: Imagenet classification using binary convolutional neural
  networks.
\newblock In \emph{European Conference on Computer Vision}, pages 525--542.
  Springer.

\bibitem[{Sandler et~al.(2018)Sandler, Howard, Zhu, Zhmoginov, and
  Chen}]{mobilenet2}
Mark Sandler, Andrew Howard, Menglong Zhu, Andrey Zhmoginov, and Liang-Chieh
  Chen. 2018.
\newblock Mobilenetv2: Inverted residuals and linear bottlenecks.
\newblock \emph{arXiv preprint arXiv:1801.04381}.

\bibitem[{Sifre and Mallat(2014)}]{firstdepth}
Laurent Sifre and St{\'e}phane Mallat. 2014.
\newblock \emph{Rigid-motion scattering for image classification}.
\newblock Ph.D. thesis, Citeseer.

\bibitem[{StackOverflow(2019)}]{iosApps}
StackOverflow. 2019.
\newblock \href
  {https://www.stackoverflow.com/questions/5887248/ios-app-maximum-memory-budget}
  {ios - experiments for maximum runtime memory accesses allowed}.

\bibitem[{Stafylakis and Tzimiropoulos(2017)}]{lipres}
Themos Stafylakis and Georgios Tzimiropoulos. 2017.
\newblock Combining residual networks with lstms for lipreading.
\newblock \emph{arXiv preprint arXiv:1703.04105}.

\bibitem[{Sui et~al.(2015)Sui, Bennamoun, and Togneri}]{sui2015listening}
Chao Sui, Mohammed Bennamoun, and Roberto Togneri. 2015.
\newblock Listening with your eyes: Towards a practical visual speech
  recognition system using deep boltzmann machines.
\newblock In \emph{Proceedings of the IEEE International Conference on Computer
  Vision}, pages 154--162.

\bibitem[{Summerfield(1992)}]{summerfield1992lipreading}
Quentin Summerfield. 1992.
\newblock Lipreading and audio-visual speech perception.
\newblock \emph{Philosophical Transactions of the Royal Society of London.
  Series B: Biological Sciences}, 335(1273):71--78.

\bibitem[{Szegedy et~al.(2015)Szegedy, Liu, Jia, Sermanet, Reed, Anguelov,
  Erhan, Vanhoucke, and Rabinovich}]{seconddepth}
Christian Szegedy, Wei Liu, Yangqing Jia, Pierre Sermanet, Scott Reed, Dragomir
  Anguelov, Dumitru Erhan, Vincent Vanhoucke, and Andrew Rabinovich. 2015.
\newblock Going deeper with convolutions.
\newblock In \emph{Proceedings of the IEEE conference on computer vision and
  pattern recognition}, pages 1--9.

\bibitem[{Vasilache et~al.(2014)Vasilache, Johnson, Mathieu, Chintala,
  Piantino, and LeCun}]{fft2}
Nicolas Vasilache, Jeff Johnson, Michael Mathieu, Soumith Chintala, Serkan
  Piantino, and Yann LeCun. 2014.
\newblock Fast convolutional nets with fbfft: A gpu performance evaluation.
\newblock \emph{arXiv preprint arXiv:1412.7580}.

\bibitem[{van~der Walt et~al.(2011)van~der Walt, Chris~Colbert, and
  Varoquaux}]{numpy}
StÃ©fan van~der Walt, S~Chris~Colbert, and Gael Varoquaux. 2011.
\newblock \href {https://doi.org/10.1109/MCSE.2011.37} {The numpy array: A
  structure for efficient numerical computation}.
\newblock \emph{Computing in Science \& Engineering}, 13:22 -- 30.

\bibitem[{Wand et~al.(2016)Wand, Koutn{\'\i}k, and
  Schmidhuber}]{wand2016lipreading}
Michael Wand, Jan Koutn{\'\i}k, and J{\"u}rgen Schmidhuber. 2016.
\newblock Lipreading with long short-term memory.
\newblock In \emph{2016 IEEE International Conference on Acoustics, Speech and
  Signal Processing (ICASSP)}, pages 6115--6119. IEEE.

\bibitem[{Wu et~al.(2018)Wu, Wan, Yue, Jin, Zhao, Golmant, Gholaminejad,
  Gonzalez, and Keutzer}]{shift}
Bichen Wu, Alvin Wan, Xiangyu Yue, Peter Jin, Sicheng Zhao, Noah Golmant, Amir
  Gholaminejad, Joseph Gonzalez, and Kurt Keutzer. 2018.
\newblock Shift: A zero flop, zero parameter alternative to spatial
  convolutions.
\newblock In \emph{Proceedings of the IEEE Conference on Computer Vision and
  Pattern Recognition}, pages 9127--9135.

\bibitem[{Wu et~al.(2016)Wu, Leng, Wang, Hu, and Cheng}]{quantization2}
Jiaxiang Wu, Cong Leng, Yuhang Wang, Qinghao Hu, and Jian Cheng. 2016.
\newblock Quantized convolutional neural networks for mobile devices.
\newblock In \emph{Proceedings of the IEEE Conference on Computer Vision and
  Pattern Recognition}, pages 4820--4828.

\bibitem[{Ye et~al.(2018)Ye, Liu, and Zhang}]{depthwise3d}
Rongtian Ye, Fangyu Liu, and Liqiang Zhang. 2018.
\newblock 3d depthwise convolution: Reducing model parameters in 3d vision
  tasks.
\newblock \emph{arXiv preprint arXiv:1808.01556}.

\bibitem[{Yu et~al.(2017)Yu, Liu, Wang, and Tao}]{tensorfactor}
Xiyu Yu, Tongliang Liu, Xinchao Wang, and Dacheng Tao. 2017.
\newblock On compressing deep models by low rank and sparse decomposition.
\newblock In \emph{Proceedings of the IEEE Conference on Computer Vision and
  Pattern Recognition}, pages 7370--7379.

\bibitem[{Zhang et~al.(2018{\natexlab{a}})Zhang, Wang, Li, and
  Wang}]{zhang2018dynamically}
Jie Zhang, Xiaolong Wang, Dawei Li, and Yalin Wang. 2018{\natexlab{a}}.
\newblock Dynamically hierarchy revolution: dirnet for compressing recurrent
  neural network on mobile devices.
\newblock \emph{arXiv preprint arXiv:1806.01248}.

\bibitem[{Zhang et~al.(2018{\natexlab{b}})Zhang, Zhou, Lin, and Sun}]{shuffle}
Xiangyu Zhang, Xinyu Zhou, Mengxiao Lin, and Jian Sun. 2018{\natexlab{b}}.
\newblock Shufflenet: An extremely efficient convolutional neural network for
  mobile devices.
\newblock In \emph{The IEEE Conference on Computer Vision and Pattern
  Recognition (CVPR)}.

\end{thebibliography}
\bibliographystyle{acl_natbib}

\clearpage
\newpage
\pagebreak
\appendix
\section{Appendices}
\label{sec:appendix}
\subsection{Calculating memory access}
\label{memory_access}
In order to estimate memory access we make some simplifying assumptions. First, to get a fair comparison we neglect the effect of computer architecture dependent optimizations. Second, for comparing memory usage by different models, we focus on two aspects that lead to memory consumption: i) memory read operations for model parameters, and ii) the memory read by a layer to read input and write its output. Third, we assume that during the forward pass, model weights are read once and then used as long as required in one pass. Each read and write is counted as one memory access. This is not necessarily true in all cases as some computing environments can read more than one value in a single memory access \cite{intel8086}. For the sake of simplicity we ignore such architectural characteristics.

\noindent \textbf{Memory access in convolutions} -
\label{section:convmem}
%For a normal two dimensional convolution of kernel size $K \times K$, having input channels $C_i$ and $C_o$ output channels, the number of parameters is 
%\begin{equation}P_{conv2d} = K^2C_iC_o \end{equation} Similarly, 
For a 3D-kernel, with ${\scriptstyle T}$ the kernel size along the temporal dimension, the number of
parameters is given by %\begin{equation} 
 ${\scriptstyle P_{conv3d} = K^2TC_iC_o}$.
% \end{equation}
Depthwise separable convolutions can be thought of as a two step process. The first part consists of convolving each channel separately which is then followed by a pointwise convolution across the full channel length of the input.
As mentioned in \cite{mobilenet1}, the number of parameters in a two dimensional depthwise separable convolution is given by
% \begin{equation}
${\scriptstyle P_{depth2d} = C_i \cdot(K^2 + C_o)}$.
% \end{equation}
The first term is the cost of the first step in a depthwise separable kernel and the second term is the cost of applying ${\scriptstyle C_o}$ number of pointwise kernels as part of the second step.
By the same logic, generalising the number of parameters to a three dimensional depthwise separable convolution layer, we get:
\begin{equation}
\scriptstyle P_{depth3d} = C_i \cdot(K^2 + C_o) \cdot(T)
\end{equation}

\noindent To calculate the number of memory reads, consider the following case. If the input has dimensions ${\scriptstyle I \times I \times C_i}$ (where ${\scriptstyle I \times I}$ is the height and width of the input matrix and ${\scriptstyle C_i}$ is the channel length), then for each convolution in two dimensions with a ${\scriptstyle K \times K}$ kernel, each element of the input matrix will be loaded from memory ${\scriptstyle K \times K}$ times. Since there are ${\scriptstyle C_o}$ number of such kernels, each element will be read ${\scriptstyle K \times K \times C_o}$ times. So the number of times input memory read operations will be performed is given by
\begin{equation} \label{equation:readconv2d1}
\scriptstyle R_{conv2d} = (I^2C_i) \cdot(K^2C_o)
\end{equation}
Here ${\scriptstyle I^2C_i}$ is the input volume ${\scriptstyle V_i}$. So we can rewrite Eq. \ref{equation:readconv2d1} as
% \begin{equation} \label{equation:readconv2d2}
${\scriptstyle R_{conv2d} = V_i\cdot(K^2C_o)}$.
% \end{equation}
Similarly for each three dimensional convolution the number of memory read operations is
% \begin{equation}
${\scriptstyle R_{conv3d} = (I^2L_iC_i)\cdot(K^2C_o) \cdot T}$.
% \end{equation}
Putting ${\scriptstyle V_i = I^2L_iC_i}$ where ${\scriptstyle V_i}$ is the input volume to 3D convolution, we get
\begin{equation}
\scriptstyle R_{conv3d} = V_i\cdot(K^2C_o) \cdot T
\end{equation}

\noindent In case of two dimensional depthwise separable convolutions, each element of the input matrix is convolved ${\scriptstyle K \times K \times 1}$ in the first step. Since each input channel has a separate spatial kernel in this step, the number of memory reads turn out to be ${\scriptstyle I \times I \times K \times K \times 1 \times C_i}$. The resultant matrix then undergoes pointwise convolution which requires ${\scriptstyle I \times I \times C_i \times 1 \times 1 \times C_o}$ memory reads. Therefore the total number of input read operations in performing a two dimensional depthwise separable convolution is 

\begin{equation} 
\scriptstyle R_{depth2d} = I^2 \times K^2 \times 1 \times C_i + I^2 \times C_i \times 1^2 \times C_o
= (I^2C_i)\cdot(K^2+C_o)
\end{equation}

which can be written as
\begin{equation}
\scriptstyle R_{depth2d} = V_i\cdot(K^2+C_o)
\end{equation}
Extending the argument to three dimensions, the number of input memory read associated with a 3D depthwise convolution layer is 
\begin{equation}
\scriptstyle R_{depth3d} = (I^2C_iL_i)\cdot(K^2 + C_o) \cdot T
\end{equation}
which is the same as
\begin{equation}
\scriptstyle R_{depth3d} = V_i \cdot(K^2 + C_o)\cdot T
\end{equation}

\noindent Memory accessed due to memory write operations for storing the output of the convolution is simply equal to the size of the output ${\scriptstyle V_o}$. Therefore the total memory access ${\scriptstyle M_{acc}}$ in convolution layers can be obtained by adding the respective ${\scriptstyle Rs, Ps}$ and ${\scriptstyle V_{o}}$.

\noindent \textbf{Memory access in fully connected layers} -
A fully connected layer which takes an ${\scriptstyle I}$ element vector and outputs a ${\scriptstyle Q}$ element vector is essentially a ${\scriptstyle I \times Q}$ weight matrix. Therefore the number of parameters in a fully connected layer is equal to the size of this matrix
%\begin{equation}
${\scriptstyle P_{fc} = IQ }$.
%\end{equation}
Since this layer is just a matrix multiplication, the input matrix needs to be read once. Hence the number of input memory read operations is simply equal to the number of elements in the input. Similar to convolution layers, the number of memory write operations in a fully connected layer is equal to the size of the output.

\subsection{Calculation of FLOPs}
To calculate the number of FLOPs in convolutions and fully connected layers, it is important to note that these operations involve element-wise multiplication (of the kernel elements with a specific region of the input feature map) followed by an addition (accumulation) of all the products obtained. For instance a dot product of two ${\scriptstyle n}$ element vectors has ${\scriptstyle n}$ element-element multiplications and ${\scriptstyle n-1}$ additions. The total number of FLOPs in an ${\scriptstyle n}$ element dot product is ${\scriptstyle 2n-1}$. As ${\scriptstyle n \gg 1}$, this could be approximated as ${\scriptstyle 2n}$.

\noindent \textbf{FLOPs in Convolutions} \label{section:convflops} -
 In the case of a two dimensional convolution, we can think of the process of convolving over a region as the dot product between the kernel weights and the input region below it. This dot-product has ${\scriptstyle K \times K \times C_i}$ elements. Therefore this operation requires ${\scriptstyle 2 \times K \times K \times C_i}$ FLOPs. This process happens for every element of the output feature map of size ${\scriptstyle H \times H}$, repeated for ${\scriptstyle C_o}$ convolution kernels. Therefore the total number of FLOPs is 
\begin{equation}
\scriptstyle F_{conv2d} = 2K^2C_iH^2C_o
\end{equation}
Here ${\scriptstyle H^2C_o}$ is simply the output volume ${\scriptstyle V_o}$.
Therefore
% \begin{equation}
${\scriptstyle F_{conv2d} = 2(K^2C_i) \cdot V_o}$.
% \end{equation}
Similarly in three dimensions this number is
% \begin{equation}
${\scriptstyle F_{conv3d} = 2H^2L_oC_iK^2TC_o}$.
% \end{equation}
As ${\scriptstyle V_o = H^2L_oC_o}$, we can write the above equation as
\begin{equation}
\scriptstyle F_{conv3d} = 2(K^2TC_i)\cdot V_o
\end{equation}
The first step of a depthwise separable convolution is similar to a normal convolution, except the dot products involve ${\scriptstyle K \times K \times 1}$ elements along each one of the ${\scriptstyle C_i}$ input channels. Similar to simple convolutions, this is done for each ${\scriptstyle H \times H}$ output element. This results in ${\scriptstyle 2 \times K^2 \times C_i \times H \times H}$ FLOPs. 

The pointwise convolution step involves dot products of ${\scriptstyle C_i}$ done for every ${\scriptstyle H \times H}$ output pixel. As there are ${\scriptstyle C_o}$ of these pointwise kernels the FLOPs required in this step are ${\scriptstyle 2 \times 1^2 \times C_i \times H^2 \times C_o}$. Therefore the total number of FLOPs is 
\begin{equation}
\scriptstyle F_{depth2d} = 2(H^2C_i)\cdot(K^2+C_o) 
\end{equation}
Again putting ${\scriptstyle V_o = H^2C_o}$ we get
% \begin{equation}
${\scriptstyle F_{depth2d} = 2C_i \cdot(\frac{K^2}{C_o}+1) \cdot V_o}$.
% \end{equation}
By the same logic, the number of FLOPs in a three dimensional depthwise convolution layers is given by
% \begin{equation}
${\scriptstyle F_{depth3d} = 2(H^2L_oC_i)\cdot(K^2+C_o) \cdot T}$,
% \end{equation}
which is the same as
\begin{equation}
\scriptstyle F_{depth3d} = 2C_i\cdot(\frac{K^2}{C_o}+1)\cdot T\cdot V_o
\end{equation}

\noindent \textbf{FLOPs in Fully Connected Layers} -
Since a fully connected layer is a ${\scriptstyle I \times Q}$ matrix where $I$ is the size of the input vector, the matrix multiply as part of the fully connected layer can be thought of as a dot product between the ${\scriptstyle I}$ element input and column vectors of the matrix which is repeated ${\scriptstyle Q}$ times. Therefore the number of FLOPs in a fully connected layer is
%\begin{equation}
${\scriptstyle F_{fc} = 2IQ }$.
%\end{equation}

Table \ref{table:memaccflop}, summarizes all the equations used for calculating memory accesses and number of FLOPs by different layers in our proposed architecture. As an example a 2D convolution layer with kernel size $3 \time 3$ with 64 as output channel size. Suppose the input is of the shape $100 \times 100 \times 3$. From Table \ref{table:memaccflop} the number of memory access are $(100^2 \cdot 3 \cdot 64) + (100 \cdot 100 \cdot 3)\cdot (100^2 \cdot 64) + (100^2 \cdot 64) \sim 1.94 \times 10^8$. Similarly FLOPs would be $2(100^2 \cdot 3 ) \cdot (100^2 \cdot 64) \sim 3.8 \times 10^10$.
\end{document}